\documentclass[10pt,twocolumn,letterpaper]{article}

\usepackage{iccv}
\usepackage{times}
\usepackage{epsfig}
\usepackage{graphicx}
\usepackage{amsmath}
\usepackage{amssymb}

\usepackage{float}
\usepackage{paralist}
\usepackage{amsmath}
\usepackage[linesnumbered,ruled]{algorithm2e}
\usepackage{bbding}
\usepackage{array}

\usepackage[pagebackref=true,breaklinks=true,letterpaper=true,colorlinks,bookmarks=false]{hyperref}

\usepackage{url}
\usepackage{relsize}
\usepackage[svgnames]{xcolor}
\hypersetup{
  colorlinks,
  urlcolor=Blue}

\iccvfinalcopy 

\def\iccvPaperID{1569} 

\ificcvfinal\fi
\begin{document}

\title{FCN-rLSTM: Deep Spatio-Temporal Neural Networks for \\Vehicle Counting in City Cameras}

\author{Shanghang Zhang${^\dagger}{^\ddagger}$\thanks{The first two authors contributed equally to this work.} , Guanhang Wu${^\dagger}{^*}$, Jo\~{a}o P. Costeira$^\ddagger$, Jos\'{e} M. F. Moura$^\dagger$\\$^\dagger$Carnegie Mellon University, Pittsburgh, PA, USA\\$^\ddagger$ISR - IST, Universidade de Lisboa, Lisboa, Portugal\\  {\small\texttt{\{shanghaz, guanhanw\}@andrew.cmu.edu, jpc@isr.ist.utl.pt, moura@andrew.cmu.edu}}}

\maketitle

\begin{abstract}
In this paper, we develop deep spatio-temporal neural networks to sequentially count vehicles from low quality videos captured by city cameras (citycams). Citycam videos have low resolution, low frame rate, high occlusion and large perspective, making most existing methods lose their efficacy. To overcome limitations of existing methods and incorporate the temporal information of traffic video, we design a novel FCN-rLSTM network to jointly estimate vehicle density and vehicle count by connecting fully convolutional neural networks (FCN) with long short term memory networks (LSTM) in a residual learning fashion. Such design leverages the strengths of FCN for pixel-level prediction and the strengths of LSTM for learning complex temporal dynamics. The residual learning connection reformulates the vehicle count regression as learning residual functions with reference to the sum of densities in each frame, which significantly accelerates the training of networks. To preserve feature map resolution, we propose a Hyper-Atrous combination to integrate atrous convolution in FCN and combine feature maps of different convolution layers. FCN-rLSTM enables refined feature representation and a novel end-to-end trainable mapping from pixels to vehicle count. We extensively evaluated the proposed method on different counting tasks with three datasets, with experimental results demonstrating their effectiveness and robustness. In particular, FCN-rLSTM reduces the mean absolute error (MAE) from 5.31 to 4.21 on TRANCOS; and reduces the MAE from 2.74 to 1.53 on WebCamT. Training process is accelerated by 5 times on average. 

\end{abstract}

\vspace{-0.3cm}
\section{Introduction}

\begin{figure}[]
\begin{center}
\includegraphics[width=0.45\textwidth, height = 6.3cm]{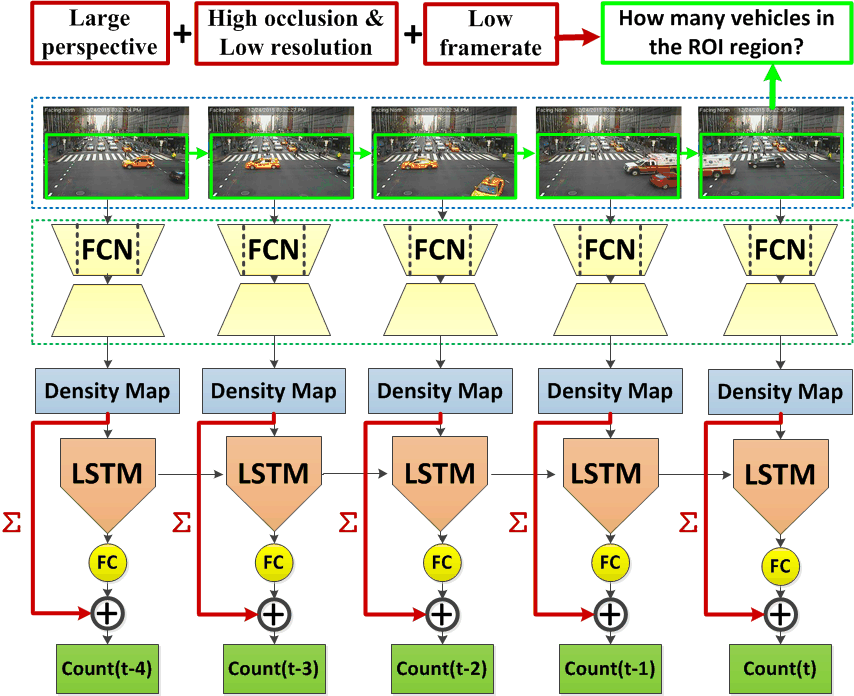}
\end{center}
\vspace{-0.2cm}
\caption{FCN-rLSTM network to count vehicles in traffic videos captured by city cameras. The videos have low frame rate, low resolution, high occlusion, and large perspective. FCN and LSTM are combined in a residual learning framework, leveraging the strengths of FCN for dense visual prediction and strength of LSTM for modeling temporal correlation. Video frames are input into FCN, and the output density maps are fed into a stack of LSTMs to learn residual functions with reference to the sum of densities in each frame. The global vehicle count is finally generated by summing the learned residual and the densities.}
\label{fig:problem}
\vspace{-0.5cm}
\end{figure}

Many cities are being instrumented with hundreds of surveillance cameras mounted on streets and intersections \cite{kastrinaki2003survey,thakurzx2012modeling,thakur2011spatial}. They capture traffic 24 hours a day, 7 days a week, generating large scale video data. Citycam videos can be regarded as highly versatile, being an untapped potential to develop many vision-based techniques for applications like traffic flow analysis and crowd counting. This paper aims to extract vehicle counts from streaming real-time video captured by citycams. Vehicle count is the number of vehicles in a given region of the road \cite{kerner2009introduction}. As shown in Figure \ref{fig:problem}, we select a region of fixed length in a video and count the number of vehicles in that region.

Vehicle counting is of great importance for many real-world applications, such as urban traffic management. Important as it is, Counting vehicles from city cameras is an extremely difficult problem faced with severe challenges (illustrated in Figure \ref{fig:problem}) due to network bandwidth limitations, lack of persistent storage, and privacy concerns. Publicly available citycam video is limited by: 
\begin{inparaenum}
\item Low frame rate, ranging from 1 fps to 0.3 fps.
\item Low resolution, including $352\times 240$, $320 \times 240$ or $704 \times 480$. 
\item High occlusion, especially in rush hours.
\item Large perspective, resulting in various vehicle scales. 
\end{inparaenum}
All these challenges make vehicle counting from citycam data very difficult.

The challenges of citycam videos preclude existing approaches to vehicle counting, which can be grouped into five categories: frame differencing based \cite{tsai2013intelligent,cucchiara2000statistic}, detection based \cite{zheng2012model,Evgeny2015Traffic}, motion based \cite{suganyadevi2012efficient,chen2011real,chen2012vehicle,mo2010vehicles}, density estimation based \cite{Lempitsky2010learning}, and deep learning based \cite{zhang2015cross, zhang2016single, onoro2016towards, zhao2016crossing, arteta2016counting} methods. Frame differencing, detection, and motion based methods are sensitive to environment conditions and tend to fail in high occlusion, low resolution, and low frame rate videos. While density estimation approaches avoid detecting or tracking individual vehicles, they perform poorly in videos with large perspective and oversized vehicles. Though the low frame rate citycam video lacks motion information, vehicle counts of sequential frames are still correlated. Existing methods fail to account for such temporal correlation \cite{onoro2016towards,webcamt,garg2016real,hua2012real,choe2010traffic,yu2002highway}. Work \cite{arteta2016counting} and \cite{webcamt} achieve state-of-the-art performance on animal counting and traffic counting, respectively, yet they fail to model the temporal correlation as an intrinsic feature of the surveillance video.

To overcome these limitations, we propose a deep spatio-temporal network architecture to sequentially estimate vehicle count by combining FCN \cite{long2015fully} with LSTM \cite{hochreiter1997long} in a residual learning framework (FCN-rLSTM). The FCN maps pixel-level features into vehicle density to avoid individual vehicle detection or tracking. LSTM layers learn complex temporal dynamics by incorporating nonlinearities into the network state updates. The residual connection of FCN and LSTM reformulates global count regression as learning residual functions with reference to the sum of densities in each frame, avoiding learning unreferenced functions and significantly accelerating the network training. FCN-rLSTM enables refined feature representation and a novel end-to-end optimizable mapping from image pixels to vehicle count. The framework is shown in Figure \ref{fig:problem}. Video frames are input into FCN, and the output density maps are fed into LSTMs to learn the vehicle count residual for each frame. The global vehicle count is finally generated by summing the learned residual and the densities.

The proposed FCN-rLSTM has the following novelties and contributions:
\begin{inparaenum}
\item FCN-rLSTM is a novel spatio-temporal network architecture for object counting, such as crowd counting \cite{zhang2015cross}, vehicle counting \cite{onoro2016towards}, and penguin counting \cite{arteta2016counting}. It leverages the strength of FCN for dense visual prediction and the strengths of LSTM for learning temporal dynamics. To the best of our knowledge, FCN-rLSTM is the first spatio-temporal network architecture with residual connection for object counting.

\item The residual connection between FCN and LSTM is novel and significantly accelerates the training process by 5 times on average, as shown by our experiments. Though some recent work on other visual tasks \cite{donahue2015long, wang2016cnn} also explored spatio-temporal networks, none of them combines FCN with LSTM, or has the residual connection between CNN and LSTM. FCN-rLSTM can be potentially applied to other visual tasks that both require dense prediction and exhibit temporal correlation. 　

\item One challenge for FCN based visual tasks is the reduced feature resolution \cite{chen2016deeplab} caused by the repeated max-pooling and striding. To solve this problem, we propose a Hyper-Atrous combination to integrate atrous convolution \cite{chen2016deeplab} in the FCN and to combine feature maps of different atrous convolution layers. We then add a convolution layer after the combined feature volume with $1\times1$ kernels to perform feature re-weighting. The selected features both preserve feature resolution and distinguish better foreground from background. Thus, the whole network accurately estimates vehicle density without foreground segmentation.

\item We jointly learn vehicle density and vehicle count from end-to-end trainable networks improving the accuracy of both tasks. Recent object counting literature \cite{zhang2016single, onoro2016towards, zhao2016crossing, arteta2016counting} estimates directly the object density map and sum the densities over the whole image to get the object count. But such methods suffer from large error when videos have large perspective and oversized vehicles (big bus or big truck). Our proposed multi-task framework pursues different but related objectives to achieve better local optimal, to provide more supervised information (both vehicle density and vehicle count) in the training process, and to learn better feature representation. 

\item We present comprehensive experiments on three datasets covering different counting tasks, such as vehicle counting and crowd counting, to show generalization and substantially higher accuracy of FCN-rLSTM. On the TRANCOS dataset \cite{onoro2016towards}, we improve over state-of-the-art baseline methods, reducing the MAE from 5.31 to 4.21.
\end{inparaenum}

The rest of paper is outlined as follows. Section 2 briefly reviews the related work for vehicle counting. Section 3 details the proposed FCN-rLSTM. Section 4 presents experimental results, and Section 5 concludes the paper.

\section{Related Work}
\label{sec:RelatedWorks}
In this section, we provide a brief review of related work on vehicle counting and LSTM for visual tasks.

\subsection{Vision-based Methods for vehicle counting}

Vision-based approaches deal with camera data, which have low installation costs, bring little traffic disruption during maintenance, and provide wider coverage and more detailed understanding of traffic flow \cite{nagaraj2013traffic,dangi2012image,kanungo2014smart}. They can be divided into five categories:

\begin{inparaenum}
\item \textbf{Frame differencing} methods count vehicles based on the difference between sequential frames and are easy to implement. They suffer with noise, abrupt illumination changes, and background changes \cite{tsai2013intelligent,cucchiara2000statistic}. 

\item \textbf{Detection based methods} \cite{zheng2012model,Evgeny2015Traffic} detect individual vehicles in each frame and perform poorly in low resolution and high occlusion videos. 

\item \textbf{Motion based methods} \cite{suganyadevi2012efficient,chen2011real,chen2012vehicle,mo2010vehicles} count vehicles by tracking and tend to fail with citycam videos due to their low frame rate and lack of motion information. 

\item \textbf{Density estimation based methods} deal with the limitations of detection and motion based method by mapping the dense (pixel-level) image feature into object densities, avoiding detecting or tracking each object, as shown in Figure \ref{fig:den-example}. Reference \cite{Lempitsky2010learning} casts the counting problem as estimating an image density whose integral over an image region gives the count of objects within that region. Object density is formulated as a linear transformation of each pixel feature, with a uniform weight vector applied to the whole image. This method suffers from low accuracy when the camera has large perspective and oversized vehicles occur.

\begin{figure}
\centering
\includegraphics[width=0.85\columnwidth, height = 2cm]{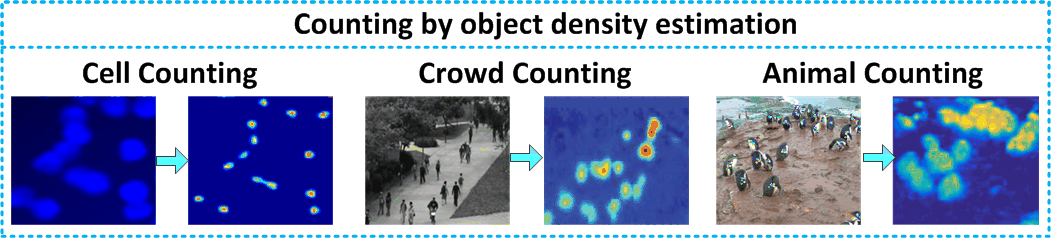}
\caption{Examples of density estimation based methods.}
\vspace{-0.3cm}
\label{fig:den-example}
\end{figure}

\item \textbf{Deep learning based counting methods} have been developed recently \cite{zhang2015cross, zhang2016single, onoro2016towards, zhao2016crossing, arteta2016counting} that significantly improved counting performance. Work \cite{zhang2015cross} applies CNN to output a 1D feature vector and fits a ridge regressor to perform the final density estimation. This work is not based on FCN and cannot perform pixel-wise dense prediction. Reference \cite{onoro2016towards} is based on FCN, but it does not have deconvolutional or upsampling layers, resulting in the output density map being much smaller than the input image. Reference \cite{arteta2016counting} jointly estimates the object density map and performs foreground segmentation, but it does not address the problem of large perspective and various object scales. 

All existing methods fail to model the temporal correlation of vehicle count in traffic video sequential frames.
\end{inparaenum}

\subsection{LSTM for Visual Tasks}

In recent years, several works attempt to combine CNN with recurrent neural networks (RNN) \cite{bahdanau2014neural} to model the spatio-temporal information of visual tasks, such as action recognition \cite{donahue2015long, baccouche2011sequential}, video description \cite{donahue2015long}, caption generation \cite{sohlearning}, and multi-label classification \cite{wang2016cnn}. However, no existing work models the spatio-temporal correlation for object counting, especially by combining CNN/FCN with RNN/LSTM. Some work \cite{neil2016phased} explores new design of the internal LSTM architecture, but none of the existing works combined FCN with LSTM in a residual learning fashion. Work \cite{zhang2016deep} regards the crowd flow map in a city as an image and build spatio-temporal networks to predict crowd flow. It does not apply RNN or LSTM networks to learn the temporal information; instead, it aggregates the output of three residual neural networks to model temporal dynamics. Thus such work is essentially multiple convolutional neural networks, rather than the combination of CNN and LSTM.  

\begin{figure*}
\begin{center}
\includegraphics[width=1.9\columnwidth, height = 4cm]{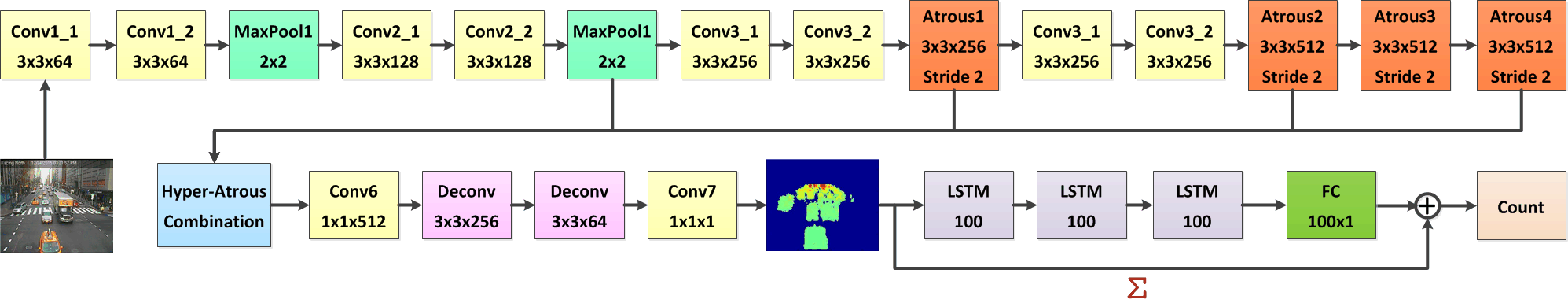}
\end{center}
\vspace{-0.3cm}
\caption{Network architecture and parameters of FCN-rLSTM.}
\vspace{-0.3cm}
\label{fig:arch}
\end{figure*}

\section{FCN-rLSTM for vehicle counting}

As the low spatial and temporal resolution and high occlusion of citycam videos preclude existing detection or motion based methods for vehicle counting, we propose to apply FCN \cite{long2015fully} to map the dense (pixel-level) feature into vehicle density and to avoid detecting or tracking individual vehicles. FCN based density estimation allows arbitrary input resolution and outputs vehicle density maps that are of the same size as the input image. Existing object counting literature \cite{zhang2016single, onoro2016towards, zhao2016crossing, arteta2016counting} estimates the object density map and directly sums the density over the whole image to get the object count. But such methods suffer from large error when the video has large perspective and oversized vehicles (big bus or big truck). Thus we propose the FCN-rLSTM network to jointly estimate vehicle density and vehicle count by connecting FCN with LSTM in a residual learning fashion. Such design leverages the strengths of FCN for pixel-level prediction and the strengths of LSTM for learning complex temporal dynamics. Counting accuracy is significantly improved by taking the temporal correlation of vehicle counts into account. However, it is not easy to train the combined FCN and LSTM networks. We further propose the residual connection of FCN and LSTM to accelerate the training process. The resulting end-to-end trainable network has high convergence rate and further improves the counting accuracy. In the following subsections, we will explain the proposed network architecture and highlight additional details.

\subsection{FCN-rLSTM Model \& Network Architecture}
The network architecture with detailed parameters is shown in Figure \ref{fig:arch}, which contains convolution network, deconvolution network, hyper-atrous feature combination, and LSTM layers. Inspired by the VGG-net \cite{simonyan2014very}, small kernels of size $3\times 3$ are applied to both convolution layers and deconvolution layers. The number of filter channels in the higher layers are increased to compensate for the loss of spatial information caused by max pooling. 

\begin{figure}[H]
\begin{center}
\includegraphics[width= 0.6\columnwidth, height = 3cm]{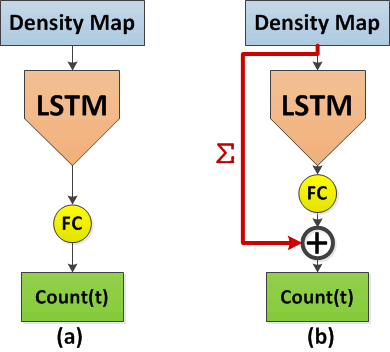}
\end{center}
\vspace{-0.3cm}
\caption{Comparison of (a) Direct connection of FCN and LSTM (FCN-dLSTM); (b) Residual connection of FCN and LSTM.}
\vspace{-0.3cm}
\label{fig:RLSTM}
\end{figure}

To preserve feature map resolution, we develop hyper-atrous combination, where atrous convolution \cite{chen2016deeplab} is integrated into the convolution networks, and the feature maps after the second max-pooling layer and the atrous convolution layers are combined together into a deeper feature volume. Atrous convolution is proposed by work \cite{chen2016deeplab}; it amounts to filter upsampling by inserting holes between nonzero filter taps. It computes feature maps more densely, followed by simple bilinear interpolation of the feature responses back to the original image size. Compared to regular convolution, atrous convolution effectively enlarges the field of view of filters without increasing the number of parameters. After several atrous convolution layers, we combine the features from the second max-pooling layer and the atrous convolution layers. And then, after the combined feature volume, we add a convolution layer with $1\times1$ kernels to perform feature re-weighting to encourage the re-weighted feature volume to distinguish better foreground and background pixels. The combined and re-weighted feature volume is input of the deconvolution network that contains two deconvolution layers. At the top of the FCN, a convolution layer with $1\times1$ kernel acts as a regressor to map the features into vehicle density.
\begin{equation}
\begin{aligned}
i_{t} & = \sigma _{i}(x_t W_{xi} + h_{t-1}W_{hi} + w_{ci}\odot c_{t-1} +b_i) \\
f_{t} & = \sigma _{f}(x_t W_{xf} + h_{t-1}W_{hf} + w_{cf}\odot c_{t-1} +b_f) \\
c_{t} & = f_t\odot c_{t-1} + i_t \odot  \sigma _{c}(x_t W_{xc} + h_{t-1}W_{hc} + b_c) \\
o_{t} & = \sigma _{o}(x_t W_{xo} + h_{t-1}W_{ho} + w_{co}\odot c_t + b_o) \\
h_{t} & = \sigma _{t}\odot \sigma _{h}(c_t)
\end{aligned}
\end{equation}
To incorporate the temporal correlation of vehicle counts from sequential frames, we combine LSTM with FCN to jointly learn vehicle density and count. RNN maintains internal hidden states to model the dynamic temporal behavior of sequences. LSTM extends RNN by adding three gates to an RNN neuron: a forget gate $f_t$; an input gate $i_t$; and an output gate $o_t$. These gates enable LSTM to learn long-term dependencies in a sequence, and make it easier to be optimized. LSTM effectively deals with the gradient vanishing/exploding issues that commonly appear during RNN training \cite{pascanu2013difficulty}. It also contains cell activation vector $c_t$ and hidden output vector $h_t$. We reshape the output density map of FCN into a 1D vector $x_t$ and feed this vector into three LSTM layers. Each LSTM layer has 100 hidden units and is unrolled for a window of 5 frames. The gates apply sigmoid nonlinearities $\sigma _{i}$, $\sigma _{f}$, $\sigma _{o}$, and tanh nonlinearities $\sigma _{c}$, and $\sigma _{h}$ with weight parameters $W_{hi}$, $W_{hf}$, $W_{ho}$, $W_{xi}$, $W_{xf}$, and $W_{xo}$, which connect different inputs and gates with the memory cells, outputs, and biases $b_i$, $b_f$, and $b_o$. We define the commonly-used update equations \cite{graves2013generating}:

To accelerate training, FCN and LSTM are connected in a residual learning fashion as illustrated in Figure \ref{fig:RLSTM}. We take the sum of the learned density map over each frame as a base count, and feed the output hidden vector of the last LSTM layer into one fully connected layer to learn the residual between base count and final estimated count. Compared to the direct connection of FCN and LSTM, the residual connection eases the training process and increases counting accuracy.

\subsection{Spatio-Temporal Multi-Task Learning}

The ground truth supervision for FCN-rLSTM includes two types of information: the pixel-level density map and the global vehicle count for each frame. Generation of these supervision depends on how the objects are annotated. If the center of each object is labeled as a dot $d$, the ground truth vehicle count for frame $i$ is the total number of labeled dots. The ground truth density $F^0_i(p)$ for each pixel $p$ in image $i$ is defined as the sum of 2D Gaussian kernels centered at each dot annotation covering pixel $p$:
\begin{equation}
F^0_i(p) = \sum_{d\in D_i} N(p; d, \delta)
\end{equation}
where $D_i$ is the set of the dot annotations, $d$ is each annotation dot, and $\delta $ of the Gaussian kernel is decided by the perspective map. If each object is annotated by a bounding box $B = (x_1, y_1, x_2, y_2)$, where $(x_1, y_1)$ are the coordinates of the left top point and $(x_2, y_2)$ are the coordinates of the right bottom point, the ground truth vehicle count for frame $i$ is the total number of bounding boxes in frame $i$. The center $o$ of each bounding box $B$ is: $o_x =\frac{1}{2}(x_1+x_2), o_y = \frac{1}{2}(y_1+y_2)$. Then, the ground truth density $F^0_i(p)$ for each pixel $p$ in image $i$ is defined as:
\begin{equation}
F^0_i(p) =\sum_{o\in O_i} N(p; o, \delta)
\end{equation}
where the parameter $O_i$ is the set of bounding box centers in frame i. $\delta $ of the Gaussian kernel is decided by the length of the bounding box.

The FCN task is to estimate the pixel-level density map, and the LSTM task is to estimate the global vehicle count for each frame. These two tasks are jointly achieved by training the whole FCN-rLSTM network end-to-end. The vehicle density is predicted from the feature map by the last convolution $1\times1$ layer of the FCN. Euclidean distance is adopted to measure the difference between the estimated density and the ground truth. The loss function for density map estimation is defined as follows:
\begin{equation}
L_{D} =\frac{1}{2N}\sum_{i = 1}^{N}\sum_{p=1}^{P}  \left \|  F_i(p; \Theta ) - F^0_{i}(p))\right \|_{2}^{2}
\end{equation}
where $N$ is the batch size and $F_{i}(p)$ is the estimated vehicle density for pixel $p$ in image $i$, and $\Theta$ is the parameter of FCN. The second task, global count regression, is learned from the LSTM layers including two parts: (i) base count: the integration of the density map over the whole image; (ii) residual count: learned by the LSTM layers. We sum the two to get the estimated vehicle count:
\begin{equation}
C_i= G(F_i;\Gamma, \Phi) + \sum_{p=1}^{P}F_i(p)
\end{equation}
where $G(F_i;\Gamma, \Phi)$ is the estimated residual count, $F_i$ is the estimated density map for frame i, $\Gamma$ is the learnable parameters of LSTM, and $\Phi$ is the learnable parameters of the fully connected layers. We hypothesize that it is easier to optimize the residual mapping than to optimize the original mapping. The loss of the global count estimation is:
\begin{equation}
L_C = \frac{1}{2N} \sum_{i=1}^{N} (C_i-C^0_i)^{2}
\end{equation}
where $C^0_i$ is the ground truth vehicle count of frame $i$, $C_i$ is the estimated count of frame $i$. Then overall loss function for the network is defined as:
\begin{equation}
L= L_{D} + \lambda L_C
\end{equation}
where $\lambda$ is the weight of the global count loss, and it should be tuned to achieve best accuracy. By simultaneously learning the two related tasks, each task can be better trained with much fewer parameters. 

The loss function is optimized via batch-based Adam \cite{kingma2014adam} and backpropagation. Algorithm \ref{alg:FCN-rLSTM} outlines the FCN-rLSTM training process. As FCN-rLSTM can adapt to different input image resolutions and variation of vehicle scales and perspectives, it is robust to different scenes. 

\begin{algorithm}
    \caption{FCN-rLSTM Training Algorithm}\label{alg:FCN-rLSTM}
    \SetAlgoLined
    \SetKwInOut{Input}{Input}
    \SetKwInOut{Label}{Label}
    \SetKwInOut{Output}{Output}
    \Input{Images: \{${I_{11}, ..., I_{nm}}$\}, wherer $n$ is the number of sequences and $m$ is the number of unrolled frames.}
    \Label{Density Maps: \{${F^0_{11}, ..., F^0_{nm}}$\}}
    \Output{Parameters of FCN, LSTM, and FC: $\Theta, \Gamma, \Phi$}
    \For{i = 1 to max\_iteration}{
        \For{j = 1 to unroll\_number}{
            $F_{ij}$ = $\textrm{FCN}(I_{ij}; \Theta)$\\
            ${L_D}_{j}$ = $L_2(F_{ij}, F^0_{ij}$)\\
            $C_{\textrm{residual}}$ = $\textrm{FC}(\textrm{LSTM}(F_{ij}; \Gamma); \Phi)$\\
            $C_{ij}$ = $\sum F_{ij}$ + $C_{\textrm{residual}}$\\
            ${L_C}_{j}$ = $L_2(\sum F^0_{ij}, C_{ij})$\\
        }
        $L$ = $\sum {L_D}_{j}$ + $\lambda \sum {L_C}_{j}$\\
        $\Theta, \Gamma, \Phi \gets$ Adam($L$, $\Theta, \Gamma, \Phi$)
    }
\end{algorithm}

\section{Experiments}
In this session, we discuss experiments and quantitative results: \begin{inparaenum}
\item We first evaluate and compare the proposed methods with state-of-the-art methods on public dataset WebCamT \cite{webcamt}.
\item We evaluate the proposed methods on the public dataset TRANCOS \cite{onoro2016towards}. 
\item To verify the robustness and generalization of our model, we evaluate our methods on the public crowd counting dataset UCSD. \cite{chan2008privacy}.
\end{inparaenum}

\subsection{Quantitative Evaluations on WebCamT}

WebCamT is a public dataset for large-scale city camera videos, which have low resolution ($352\times240$), low frame rate (1 frame/second), and high occlusion. Both bounding box and vehicle count are available for $60,000$ frames. The dataset is divided into training and testing sets, with 45,850 and 14,150 frames, respectively, covering multiple cameras and different weather conditions.

Following the same settings in \cite{webcamt}, we evaluate our method on the 14,150 test frames of WebCamT, which contains 61 videos from 8 cameras. These videos cover different scenes, congestion states, camera perspectives, weather conditions, and time of the day. The training set contains 45,850 frames with the same resolution, but from different videos. Both training and testing sets are divided into two groups: downtown cameras and parkway cameras. Mean absolute error (MAE) is employed for evaluation. For FCN-rLSTM, the weight of the vehicle count loss is 0.01. The learning rate is initialized by 0.0001 and adjusted by the first and second order momentum in the training process. To test the efficacy of the proposed Hyper-Atrous combination, combination of FCN and LSTM, and the residual connection, we evaluate different configurations of FCN-rLSTM as shown in Table \ref{tb:tb3}. Atrous indicates the atrous convolution; Hyper indicates hypercolumn combination of the feature maps; Direct connect indicates combining FCN with LSTM directly; Residual connect indicates connecting FCN with LSTM in residual fashion.

\begin{table}[]
\small
\centering
\caption{Different configurations of FCN-rLSTM}
\vspace{0.2cm}
\label{tb:tb3}
\resizebox{0.47\textwidth}{!}{
\begin{tabular}{|c|c|c|c|c|}
\hline
Configuration & Atrous & Hyper &  \begin{tabular}{c} Direct\\ connect \end{tabular} & \begin{tabular}{c} Residual\\ connect \end{tabular} \\ \hline
FCN-A & \Checkmark & X & X & X  \\ \hline
FCN-H & X & \Checkmark & X & X    \\ \hline
FCN-HA & \Checkmark & \Checkmark & X & X    \\ \hline
FCN-dLSTM & \Checkmark & \Checkmark & \Checkmark & X   \\ \hline
FCN-rLSTM & \Checkmark & \Checkmark & X & \Checkmark   \\ \hline
\end{tabular}
}
\end{table}

\textbf{Data augmentation.}
To make the model more robust to various cameras and weather conditions, several data augmentation techniques are applied to the training images: \begin{inparaenum}
\item horizontal flip,
\item random crop,
\item random brightness,
\item and random contrast.
\end{inparaenum}
More details can be found in the released code and other data augmentation techniques can also be applied.

\textbf{Baseline approaches}. We compare our method with three methods: \emph{Baseline 1: Learning to count} \cite{Lempitsky2010learning}. This work maps each pixel's feature into object density with uniform weight for the whole image. For comparison, we extract dense SIFT features \cite{lowe2004distinctive} for each pixel using VLFeat \cite{vedaldi08vlfeat} and learn the visual words. \emph{Baseline 2: Hydra}\cite{onoro2016towards}. It learns a multi-scale non-linear regression model that uses a pyramid of image patches extracted at multiple scales to perform final density prediction. We train Hydra 3s model on the same training set as FCN-rLSTM.  \emph{Baseline 3: FCN} \cite{webcamt}. It develops a deep multi-task model to jointly estimate vehicle density and vehicle count based on FCN. We train FCN on the same training set as FCN-rLSTM.

\begin{table}[]
\centering
\caption{ Results comparison on WebCamT}
\vspace{0.2cm}
\label{tb:tb4}
\begin{tabular}{|l|l|l|}
\hline
Method & Downtown & Parkway \\ \hline
Baseline 1 & 5.91   &5.19   \\ \hline
Baseline 2 & 3.55  & 3.64   \\ \hline
Baseline 3 & 2.74  & 2.52   \\ \hline
FCN-A & 3.07   & 2.75  \\ \hline
FCN-H &2.48    & 2.30   \\ \hline
FCN-HA & 2.04  & 2.04  \\ \hline
FCN-dLSTM & 1.80  & 1.82  \\ \hline
FCN-rLSTM & \textbf{1.53}  &  \textbf{1.63} \\ \hline
\end{tabular}
\end{table}

\begin{figure}[]
\begin{center}
\includegraphics[width= 0.95\columnwidth, height = 4cm]{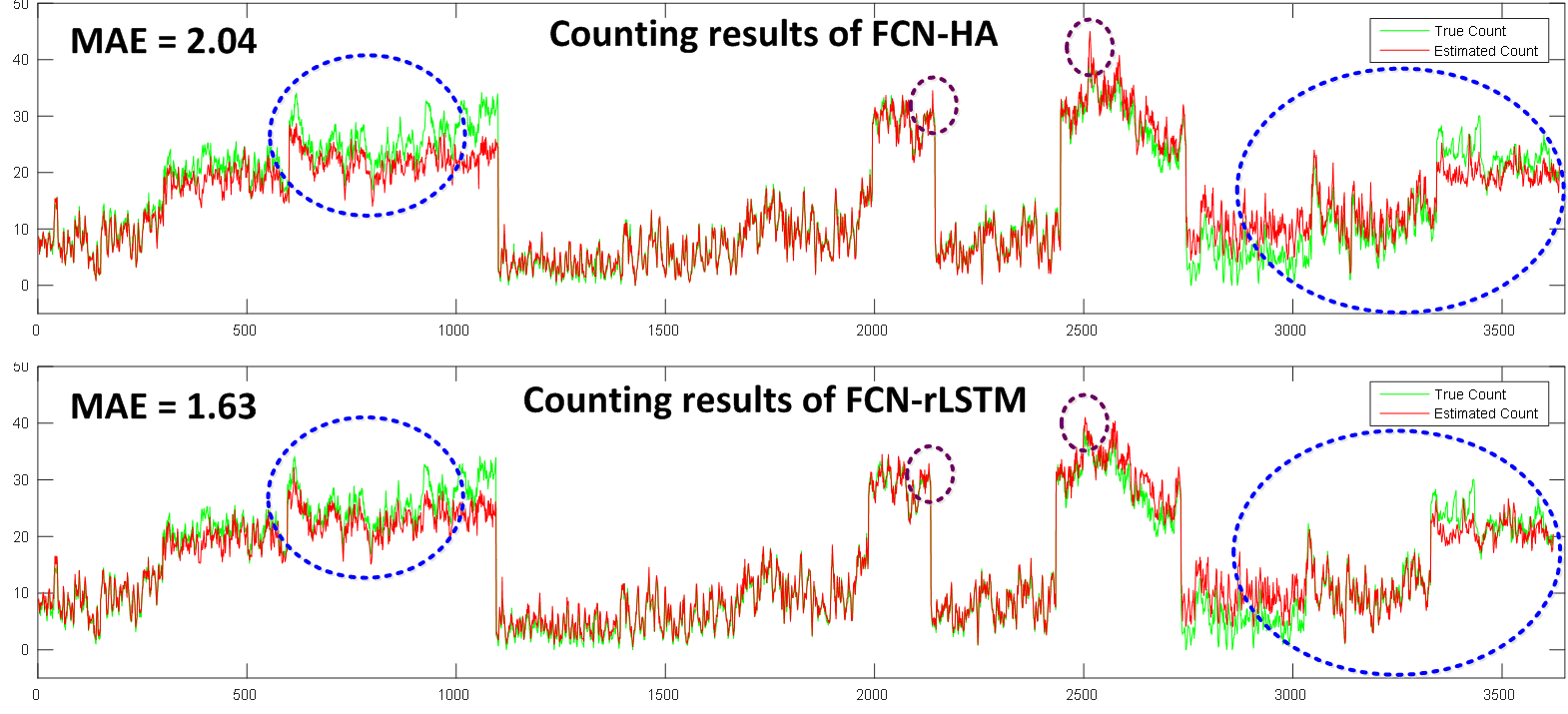}
\end{center}
\vspace{-0.3cm}
\caption{Counting results comparison of FCN-HA and FCN-rLSTM on parkway cameras. X axis-frames; Y axis-Counts.}
\vspace{-0.3cm}
\label{fig:parkway}
\end{figure}

\begin{figure}[]
\begin{center}
\includegraphics[width= 0.98\columnwidth, height = 4cm]{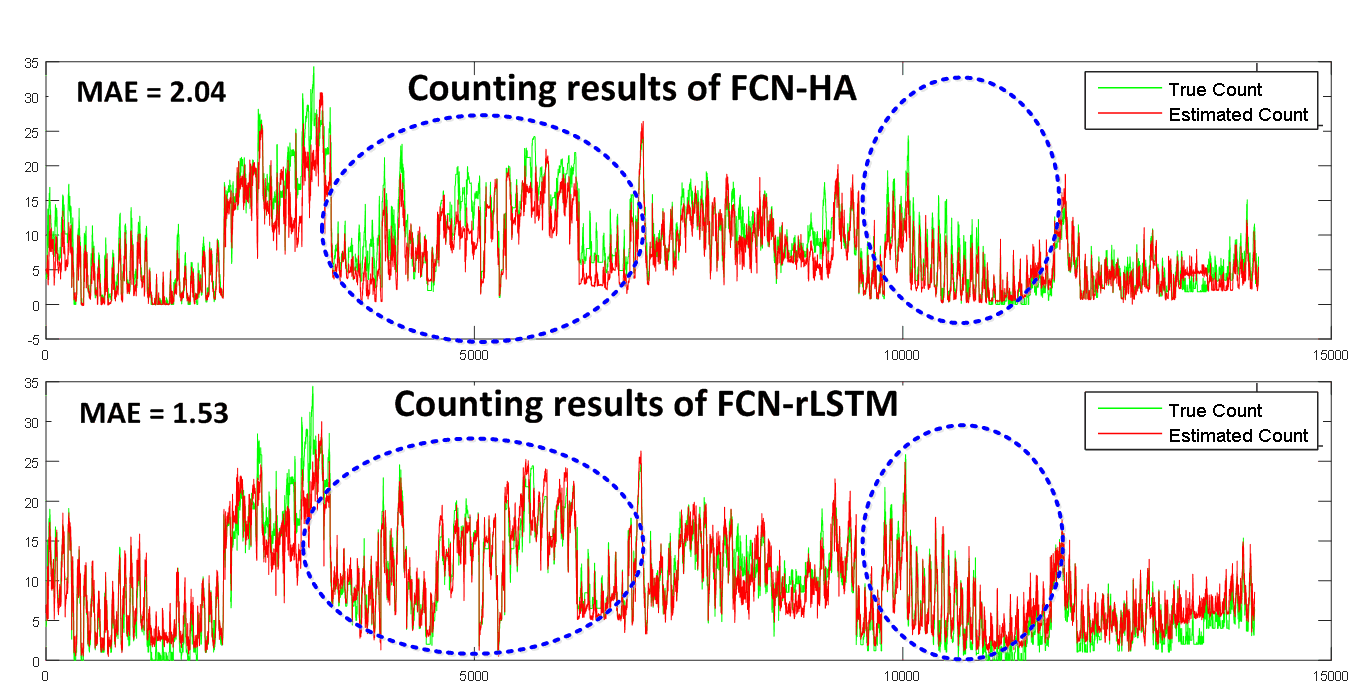}
\end{center}
\vspace{-0.3cm}
\caption{Counting results comparison of FCN-HA and FCN-rLSTM on downtown cameras. X axis-frames; Y axis-Counts.}
\vspace{-0.3cm}
\label{fig:down}
\end{figure}

\begin{figure}[H]
\begin{center}
\includegraphics[width= 0.95\columnwidth, height = 3.8cm]{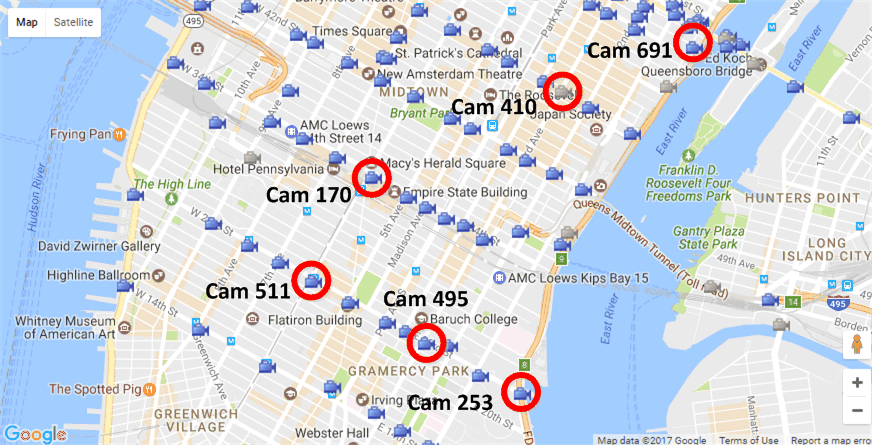}
\end{center}
\vspace{-0.3cm}
\caption{Test cameras in the urban area}
\vspace{-0.3cm}
\label{fig:map}
\end{figure}

\begin{figure*}[]
\begin{center}
\includegraphics[width= 1.85\columnwidth, height = 5.5cm]{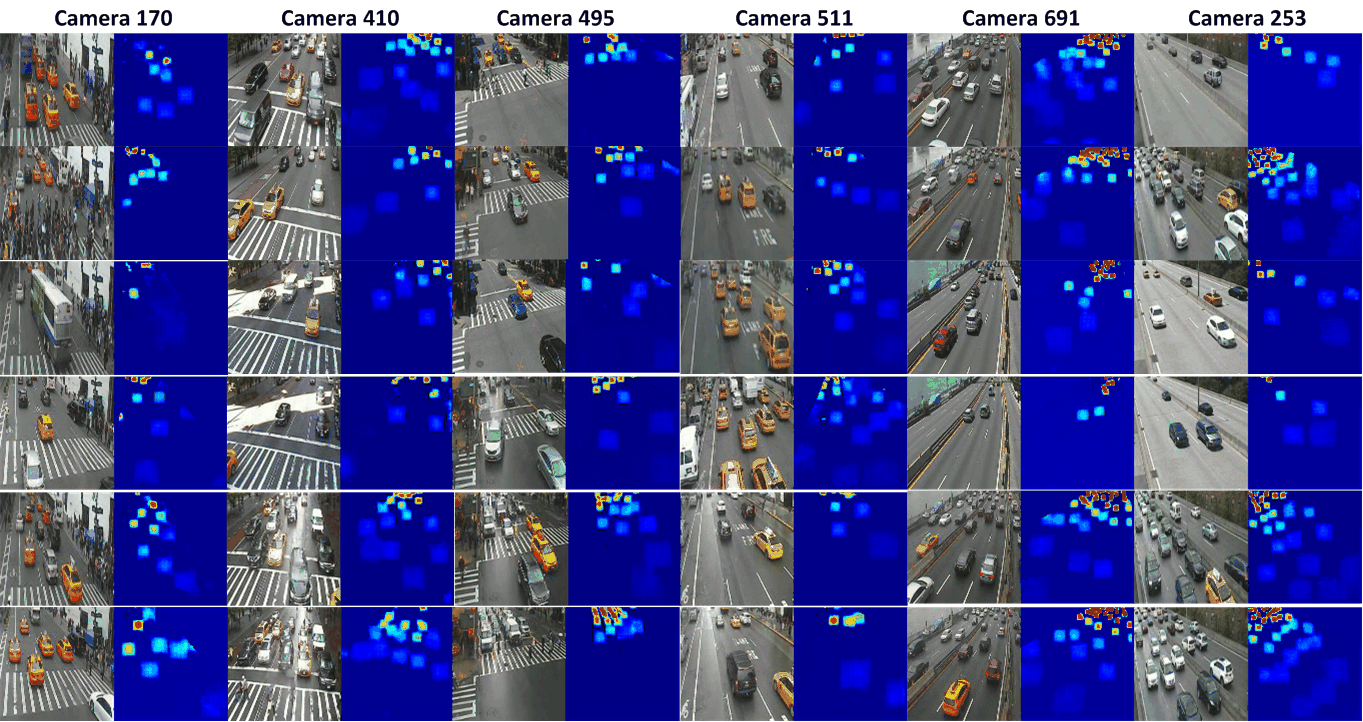}
\end{center}
\vspace{-0.3cm}
\caption{Estimated density map for multiple cameras. Column direction: The first four cameras are from downtown, and the last two cameras are from parkway. Row direction: The first two rows are estimated density maps for cloudy frames; the middle two rows are for sunny frames; and the last two rows are for rainy frames. Better view in color. Some density values may be too small to be clearly seen.}
\vspace{-0.3cm}
\label{fig:den}
\end{figure*}

\begin{figure}[H]
\begin{center}
\includegraphics[width= 0.98\columnwidth, height = 11.5cm]{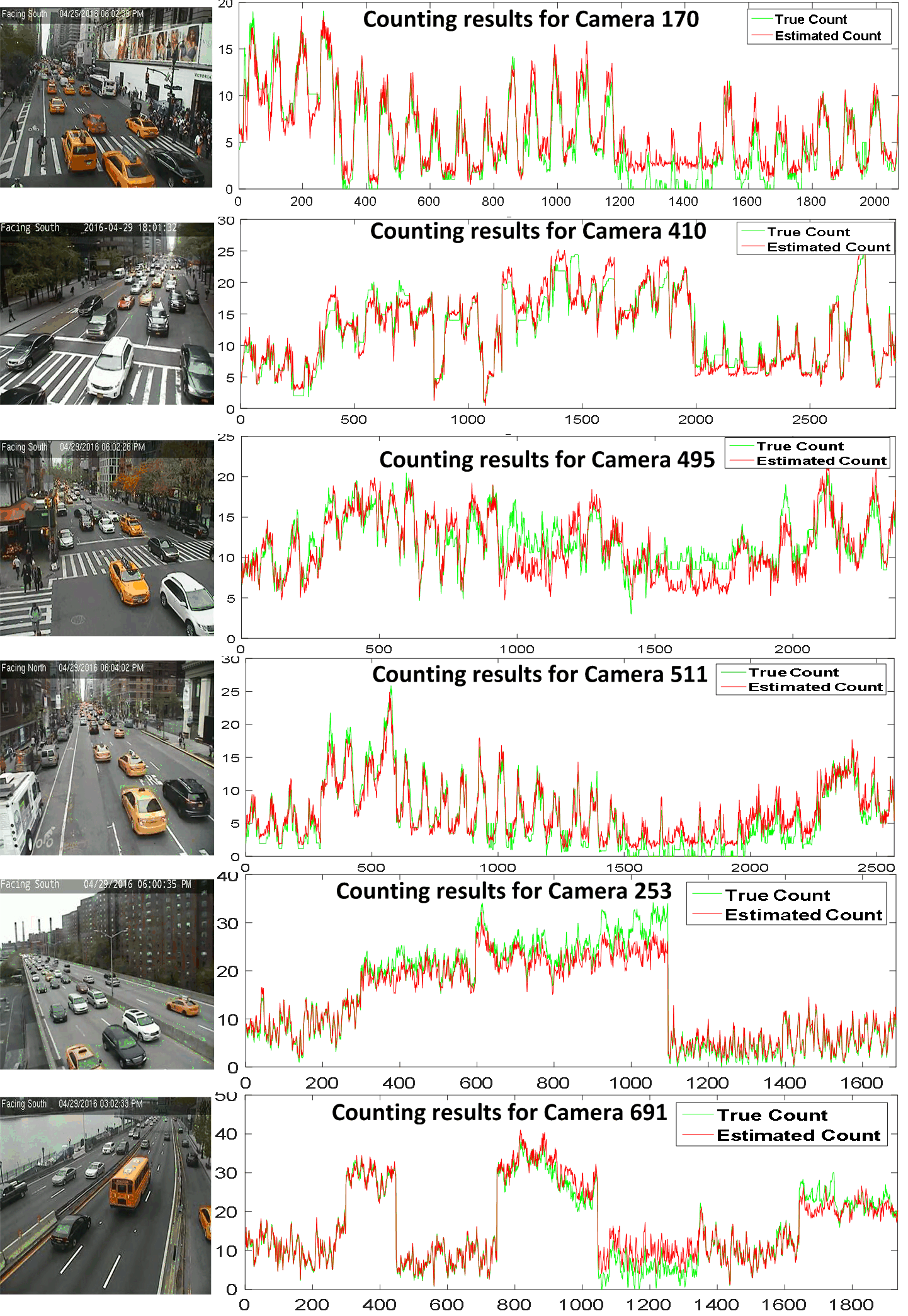}
\end{center}
\vspace{-0.3cm}
\caption{Counting results for multiple cameras.}
\vspace{-0.3cm}
\label{fig:multi-cam}
\end{figure}

\textbf{Experimental Results}. We compare the error of the proposed and baseline approaches in Table \ref{tb:tb4}. From the results, we see that FCN-rLSTM outperforms all the baseline approaches and all the other configurations. As the testing data cover different congestion states, camera perspectives, weather conditions, and time of the day, these results verify the efficacy and robustness of FCN-rLSTM. To do ablation analysis of the proposed techniques, we also evaluate the performance of different configurations as shown in Table \ref{tb:tb4}. With the Hyper-Atrous combination, FCN-HA itself already outperforms all the baseline methods and achieves better accuracy than FCN-A and FCN-H, which verifies the efficacy of the Hyper-Atrous combination. FCN-rLSTM has higher accuracy than FCN-HA and FCN-dLSTM, which verifies the efficacy of the residual connection of FCN and LSTM. Figure \ref{fig:parkway} and Figure \ref{fig:down} compare the counting results of FCN-HA and FCN-rLSTM, from which we conclude that FCN-rLSTM estimates better the vehicle count (blue dashed circles) and reduces large counting error induced by oversized vehicles (purple dashed circles).  Figure \ref{fig:den} shows the density map learned from FCN-rLSTM. Without foreground segmentation, the learned density map can still distinguish background from foreground in both sunny, rainy and cloudy, dense and sparse scenes. Figure \ref{fig:multi-cam} shows the counting results for six different cameras from downtown and parkway. The camera positions are shown in the map of Figure \ref{fig:map}. From the counting curves, we see that the proposed FCN-rLSTM accurately counts the vehicles for multiple cameras and long time sequences.

\begin{figure}[H]
\begin{center}
\includegraphics[width= 0.95\columnwidth, height = 3cm]{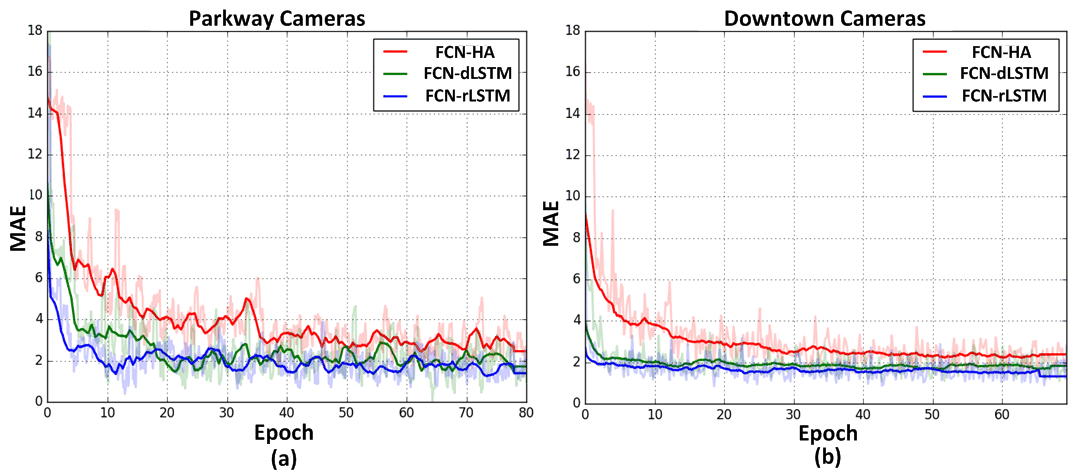}
\end{center}
\vspace{-0.3cm}
\caption{Convergence of FCN-HA, FCN-dLSTM, and FCN-rLSTM for: (a) Parkway Cameras (b) Downtown Cameras. Shading shows the MAE over Epochs and dark lines indicate the smoothed trend.}
\vspace{-0.3cm}
\label{fig:P-C}
\end{figure}

Besides the high accuracy achieved by FCN-rLSTM, the convergence of the proposed approach is also improved significantly. As shown in Figure \ref{fig:P-C}, the FCN-rLSTM converges much faster than FCN alone networks (FCN-HA). The residual connection of FCN and LSTM also enables faster convergence than the direct connection.

\subsection{Quantitative Evaluations on TRANCOS}
We also evaluate the proposed method on a public dataset TRANCOS \cite{onoro2016towards} to verify its efficacy. TRANCOS is a collection of 1244 images of different traffic scenes from surveillance camera videos. It has 46796 annotated vehicles in total and provides a region of interest (ROI) for each image. Images of TRANCOS are from very different scenarios and no perspective maps are provided. The ground truth vehicle density maps are generated by the 2D Gaussian Kernel in the center of each annotated vehicle \cite{guerrero2015extremely}.

\begin{figure}[t]
\setlength{\abovecaptionskip}{0.cm}
\setlength{\belowcaptionskip}{-0.5cm}
\begin{center}
\includegraphics[width=0.95\columnwidth, height = 7.5cm]{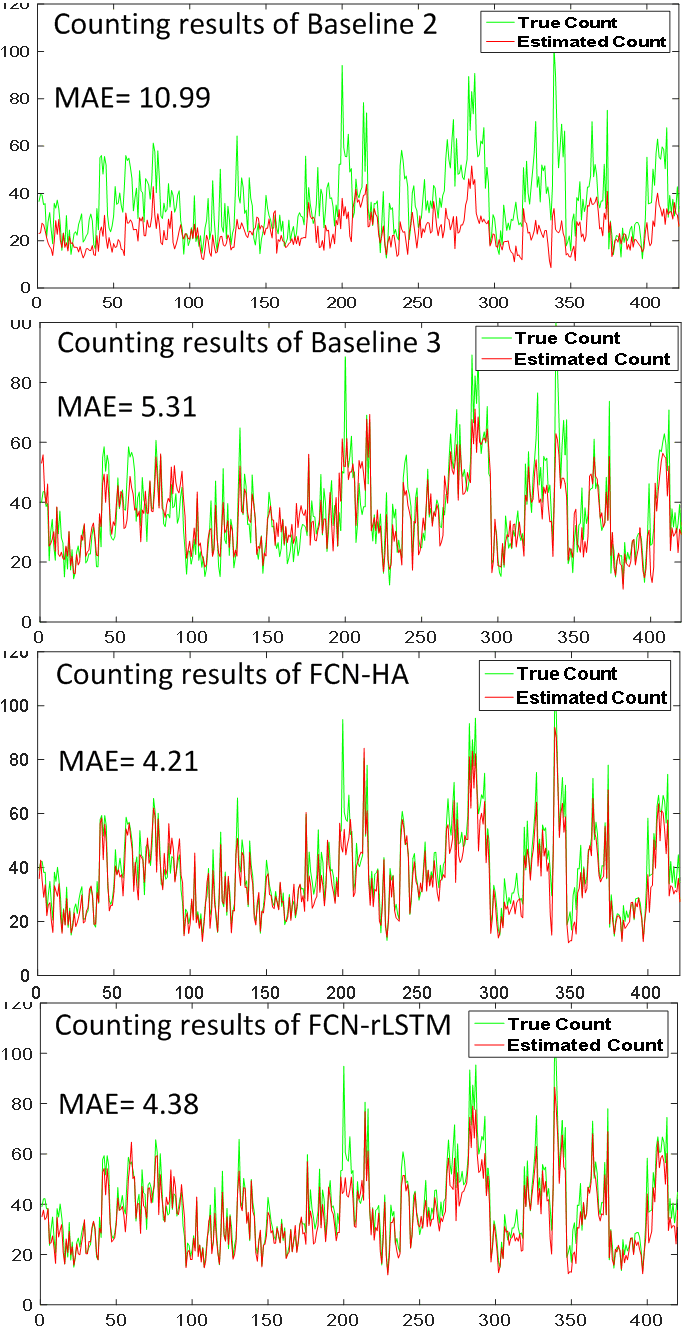}
\end{center}
\vspace{-0.3cm}
\caption{Results comparison on TRANCOS dataset.}
\label{fig:trancos}
\end{figure}

The MAE of the proposed method and baseline methods are compared in Table \ref{tb:tb5}. Baseline 2-CCNN is a basic version of the network in \cite{onoro2016towards}, and Baseline 2-Hydra augments the performance by learning a multiscale regression model with a pyramid of image patches to perform the final density prediction. All the baselines and proposed methods are trained on 823 images and tested on 421 frames following the separation in \cite{onoro2016towards}. From the results, we can see FCN-HA significantly decreases the MAE from 10.99 to 4.21 compared with Baseline 2-Hydra, and decreases the MAE from 5.31 to 4.21 compared with Baseline 3. As the training and testing images of TRANCOS are random samples from different cameras and videos, they lack consistent temporal information. Thus FCN-rLSTM cannot learn temporal patterns from the training data. The performance of FCN-rLSTM is not as good as FCN-HA, but it already outperforms all the baseline methods. When applying our proposed model to other datasets, we can choose the FCN-rLSTM configuration for datasets that have temporal correlation and choose the FCN-HA configuration for datasets that do not have temporal correlation. Figure \ref{fig:trancos} compares the estimated counts from the proposed and baseline methods. The estimated counts of the proposed methods are evidently more accurate than that of the baseline methods. FCN-rLSTM and FCN-HA have comparable estimation accuracy of vehicle counts.

\begin{table}[H]
\centering
\small
\caption{Results comparison on TRANCOS dataset}
\label{tb:tb5}
\begin{tabular}{|l|l|l|l|}
\hline
Method & MAE & Method & MAE  \\ \hline
Baseline 1 & 13.76 & Baseline 3 & 5.31\\ \hline
Baseline 2-CCNN & 12.49 & FCN-HA & \textbf{4.21}\\ \hline
Baseline 2-Hydra & 10.99 & FCN-rLSTM & 4.38 \\ \hline
\end{tabular}%
\end{table}

\setlength{\textfloatsep} {0pt plus 2pt minus 5pt}

\subsection{Quantitative Evaluations on UCSD Dataset}
To verify the generalization and robustness of our proposed methods in different counting tasks, we also evaluate and compare our methods with baselines on the pedestrian counting dataset UCSD \cite{chan2008privacy}. This dataset contains 2000 frames chosen from one surveillance camera. The frame size is $158 \times 238$ and frame rate is $10 \textrm{fps}$. Average number of people in each frame is around $25$. The dataset provides the ROI for each video frame. By following the same setting in \cite{chan2008privacy}, we use frames from 601 to 1400 as training data, and the remaining 1200 frames as test data. Table \ref{tb:tb6} shows the results of our methods and existing methods, from which we can see that FCN-rLSTM outperforms all the baseline methods and the FCN-HA configuration. These results show our proposed methods are robust to other type of counting tasks.

\begin{table}[]
\centering
\small
\caption{Results comparison on UCSD dataset}
\label{tb:tb6}
\begin{tabular}{|l|l|l|}
\hline
Method & MAE & MSE \\ \hline
Kernel Ridge Regression \cite{an2007face} & 2.16 & 7.45 \\ \hline
Ridge Regression \cite{chen2012feature} & 2.25 & 7.82 \\ \hline
Gaussian Process Regression \cite{chan2008privacy} & 2.24 & 7.97 \\ \hline
Cumulative Attribute Regression \cite{chen2013cumulative} & 2.07 & 6.86 \\ \hline
Cross-scene DNN\cite{zhang2015cross} & 1.6 & 3.31 \\ \hline
Baseline 3 & 1.67 & 3.41 \\ \hline
FCN-HA & 1.65 & 3.37 \\ \hline
FCN-rLSTM & \textbf{1.54} & \textbf{3.02} \\ \hline
\end{tabular}
\end{table}

\vspace{1cm}
\section{Discussion \& Conclusion}
Vehicle counting is of great significance for many real world applications, such as traffic management, optimal route planning, and environment quality monitoring. Counting vehicles from citycams is an extremely challenging problem as videos from citycams have low spatial and temporal resolution, and high occlusion. To overcome these challenges, we propose a novel FCN-rLSTM network architecture to jointly estimate vehicle density and vehicle count by connecting FCN with LSTM in a residual learning fashion. The residual connection reformulates global count regression as learning residual functions with reference to the sum of densities in each frame. Such design avoids learning unreferenced functions and significantly accelerates the training of the network. Extensive evaluations on different counting tasks and three datasets demonstrate the effectiveness and robustness of the proposed methods. One limitation for FCN-rLSTM is that the window size of the unrolled sequential frames is restricted by the available memory capacity. Thus we cannot learn very long term temporal information from the current FCN-rLSTM architecture. This limitation does not significantly affect the counting performance as small window sizes (five frames in this paper) is capable of learning the smoothness of vehicle count. In future work, we will input the time feature into LSTM to account for the periodicity of the city traffic flow.


{\small
\bibliographystyle{ieee}
\bibliography{egbib}
}

\end{document}